\documentclass[review]{elsarticle}

\usepackage{lineno,hyperref}
\usepackage{euscript}
\usepackage{amsthm,amsmath,amssymb,bm}
\usepackage[onehalfspacing]{setspace}
\usepackage[marginratio={1:1,1:1},scale=0.75]{geometry} 
\usepackage[ruled,vlined]{algorithm2e}
\usepackage{epstopdf} 
\usepackage{footmisc}
\usepackage{url}

\usepackage{natbib}
\usepackage{makecell}
\usepackage{multirow}
\usepackage{animate}
\usepackage{graphicx}
\usepackage{mathtools}
\DeclarePairedDelimiter\abs{\lvert}{\rvert}%
\DeclarePairedDelimiter\ceil{\lceil}{\rceil}

\journal{}

\begin{document}

\begin{frontmatter}

\title{On Integrating Prior Knowledge into Gaussian Processes \\for Prognostic Health Monitoring}

\author{Simon Pfingstl\corref{mycorrespondingauthor}}
\cortext[mycorrespondingauthor]{Corresponding author}
\ead{simon.pfingstl@tum.de}

\author{Markus Zimmermann}

\address{}
\author[mymainaddress]{Technical University of Munich, TUM School of Engineering and Design, Department of Mechanical Engineering, Laboratory for Product Development and Lightweight Design, Boltzmannstr. 15, 85748 Garching, Germany}


\begin{abstract}
Gaussian process regression is a powerful method for predicting states based on given data. It has been successfully applied for probabilistic predictions of structural systems to quantify, for example, the crack growth in mechanical structures. Typically, predefined mean and covariance functions are employed to construct the Gaussian process model. Then, the model is updated using current data during operation while prior information based on previous data is ignored. However, predefined mean and covariance functions without prior information reduce the potential of Gaussian processes.

This paper proposes a method to improve the predictive capabilities of Gaussian processes. We integrate prior knowledge by deriving the mean and covariance functions from previous data. More specifically, we first approximate previous data by a weighted sum of basis functions and then derive the mean and covariance functions directly from the estimated weight coefficients. Basis functions may be either estimated or derived from problem-specific governing equations to incorporate physical information.

The applicability and effectiveness of this approach are demonstrated for fatigue crack growth, laser degradation, and milling machine wear data. We show that well-chosen mean and covariance functions, like those based on previous data, significantly increase look-ahead time and accuracy. Using physical basis functions further improves accuracy. In addition, computation effort for training is significantly reduced.
\end{abstract}

\begin{keyword}
Gaussian processes \sep physics-informed Gaussian processes \sep prognostic health monitoring \sep fatigue damage prognosis \sep probabilistic predictions
\end{keyword}

\end{frontmatter}



\section{Introduction}

\paragraph{Motivation}
During operation, mechanical systems are constantly exposed to fatigue loads. Even if the load remains below the ultimate strength, mechanical systems can fail due to fatigue. As fatigue-induced damage is still one of the most uncertain failures in structures \cite{gobbato2014recursive} and to minimize the risk of sudden failure, mechanical systems are inspected periodically (e.g., aircraft structures see \cite{schijve2009fatigue}). However, maintenance is a significant contributor to operating costs. For example, the average direct maintenance cost of a Boeing 757-200 was 23\% of the total flight operating cost \cite{weber2017international}. Structural health monitoring (SHM), i.e., monitoring structures via data from attached sensors, enables the assessment of the structure's condition during operation \cite{farrar2010introduction} and can reduce maintenance costs. A further improvement can be obtained by predicting the condition of the mechanical systems, which is also known as prognostic health monitoring (PHM).

Using mathematical surrogate models is one way of predicting future states based on data. Researchers have proposed many different types of models for predicting fatigue life \cite{sikorska2011prognostic}. Some of them, e.g., recurrent neural networks \cite{gugulothu2017predicting,do2019fast,ma2019rotating}, support vector regression \cite{khelif2016direct,song2018predict} and k-nearest neighbors \cite{tian2014anomaly,wang2016k}, have been applied successfully to SHM and PHM problems. However, by default, these surrogate models do not provide any information about the uncertainty of the predictions.

As Gaussian process regression (GPR) is a powerful method that provides credible intervals for the predicted states, it has recently been used for probabilistic predictions \cite{mohanty2011bayesian, hong2012remaining, liu2013prognostics, kwon2015remaining, an2015practical, aye2017integrated, su2017gaussian, kong2018gaussian, yu2018state, li2019online, hassani2019physics, yuan2019pzt, avendano2020gaussian, chen2020line,  gentile2020gaussian, li2020hybrid}. A Gaussian process is completely defined by its mean and covariance function, which we also refer to as the Gaussian process model (GPM). In order to apply GPR, a mean function and a covariance function have to be chosen a priori. Table\,\ref{tb:introComp} shows mean and covariance functions that have been recently used in the field of SHM and PHM. As shown, a frequent choice is, e.g., a zero mean and a squared-exponential covariance function. However, using such predefined functions can lead to poor predictions as they involve the assumption of a specific GPM family. As stated in \cite{kwon2015remaining, kan2015review, plagemann2008nonstationary, saha2010distributed, hong2012remaining}, prior knowledge regarding the covariance function should improve the prediction of Gaussian processes.

\begin{table}[h]
\caption{Recently used mean and covariance functions for SHM and PHM problems.}
\centering
\resizebox{\textwidth}{!}{%
\begin{tabular}{l l l l l}
	\hline
	 \textbf{Topic} & \textbf{Mean function} & \textbf{Covariance function} & \textbf{Year} & \textbf{Reference} \\
	 
	 \hline

	Crack growth prediction & zero & neural network (NN) & 2011 & \citet{mohanty2011bayesian} \\

	Degradation of bearings & zero & squared-exponential (SE) iso, Matern, NN & 2012 & \citet{hong2012remaining} \\

	Battery degradation & linear, quadratic & SE iso, periodic, constant & 2013 & \citet{liu2013prognostics} \\
	
	Crack growth prediction & polynomial & SE iso & 2015 & \citet{an2015practical} \\
	
	Degradation of solder joints & zero & SE iso & 2015 & \citet{kwon2015remaining} \\
	 
	 Damage detection in bearings & \multirow{2}{3.5cm}{zero, constant, linear, quadratic, cubic} & \multirow{2}{6.5cm}{SE iso, SE ard, linear, linear ard, Matern\,iso, noise, periodic, rational-quadratic (RQ) ard, RQ iso} & 2017 & \citet{aye2017integrated} \\ \\ \\

	Battery degradation & zero & SE ard & 2018 & \citet{yu2018state} \\

	Tool wear prediction & linear & SE iso & 2018 & \citet{kong2018gaussian} \\
	
	Crack growth prediction & zero & SE iso & 2019 & \citet{li2019online} \\
	
	Degradation of bearings & linear & SE iso & 2019 & \citet{hassani2019physics} \\ 
	
	Crack growth evaluation & zero & SE iso + linear & 2019 & \citet{yuan2019pzt} \\

	Crack growth evaluation & zero & SE ard + linear ard & 2020 & \citet{chen2020line} \\

	Seismic fragility & zero & SE iso & 2020 & \citet{gentile2020gaussian} \\

	\hline
\multicolumn{5}{l}{\footnotesize{iso: isotropic length scale}} \\
\multicolumn{5}{l}{\footnotesize{ard: automatic relevance determination}} \\

\end{tabular}} \label{tb:introComp}
\end{table}

\paragraph{Contribution}
In practice, engineers usually have good a priori knowledge about the system behavior/output due to pre-executed simulations or tests. Nevertheless, defining a GPM might still be difficult, which often results in choosing default priors such as a zero mean and a squared-exponential covariance function. Therefore, we present an approach for inferring a GPM from previous data and thus integrate prior knowledge into Gaussian processes and weaken the assumption about the GPM family. Instead of implicitly making assumptions about the system output by choosing a mean and a covariance function, the user can explicitly specify which (basis) functions the system output consists of. This paves the way for physics-informed GPMs. We show that integrating this prior knowledge improves predictions in the context of PHM significantly. More specifically, the methods presented in this paper deal with low-frequency deterioration data. We use the already observed deterioration data (values and when they were observed) as inputs in order to predict the upcoming deterioration trajectory.

\paragraph{Definition of terms}
In this work, the term \textit{system output} is used for the variable that indicates the deterioration of the mechanical system. This variable should be determined in the low-frequency domain. We distinguish between \textit{previous} and \textit{current} (system output) data. The first represents all data collected by previously executed simulations or tests of a similar mechanical system. Previous data comprises several trajectories, each belonging to one simulation or test. From a statistical point of view, each trajectory represents one realization of a process. Different system outputs can be achieved by varying uncertain parameters, such as material parameters and loads. By contrast, current data describes the actual monitored system with a set of fixed, realized parameters that, in general, are unknown.

\paragraph{Structure of the paper}
After this introduction, the paper summarizes the approach of applying GPR to PHM problems based only on current data. It then presents two ways of integrating previous data into Gaussian processes. Section\,\ref{sec:poly} describes the investigated PHM problems and the application of polynomial basis functions. Furthermore, we investigate the influence of integrating previous data into Gaussian processes in this section. The effect of using physics-informed basis functions is examined in Section\,\ref{sec:physics}. In the last section, we draw conclusions from this work.


\section{Gaussian processes for PHM considering current data only} \label{sec:GPMcurr}
\noindent A Gaussian process 
\begin{equation} \label{eq:gp}
	f(x) \sim \mathcal{GP}(m_\theta(x),k_\theta(x,x'))
\end{equation}
is completely defined by its mean $m_\theta(x)$ and covariance function $k_\theta(x,x')$ which typically have some free parameters $\bm{\theta}$. A frequent choice is, for example, the squared-exponential covariance function that additionally considers noise
\begin{equation} \label{eq:se}
	k_\theta(x,x')=\sigma_f^2 \, \exp \left( -\frac{(x-x')^2}{2 \ell^2} \right) + \sigma_y^2 \mathbf{I}
\end{equation}
and a zero mean function, where $\mathbf{I}$ is the identity matrix with the same dimensionality as $x$ and $\bm{\theta}=\{\sigma_f,\ell,\sigma_y\}$ are the free parameters. Typically, these parameters are trained by maximizing the log marginal likelihood
\begin{equation} \label{eq:maximizeLogMargLike}
	\mathrm{log}\,p(\mathbf{y}_+|\mathbf{x}_+,\bm{\theta}) = -\frac{1}{2}\left(\mathbf{y}_+-m_\theta(\mathbf{x}_+)\right)^\top k_\theta(\mathbf{x}_+,\mathbf{x}_+')^{-1}\left(\mathbf{y}_+-m_\theta(\mathbf{x}_+)\right) - \frac{1}{2} \log \det k_\theta(\mathbf{x}_+,\mathbf{x}_+') - \frac{n}{2} \log\,2\pi,
\end{equation}
with $n$ as the dimensionality of the observed data, e.g., crack length, 
\begin{equation}
	\mathbf{y}_+ = \begin{bmatrix} y_{+,1} , \ldots , y_{+,n} \end{bmatrix}^\top
\end{equation}
at the locations, e.g., number of cycles, 
\begin{equation}
	\mathbf{x}_+ = \begin{bmatrix} x_{+,1} , \ldots , x_{+,n} \end{bmatrix}^\top,
\end{equation}
the mean vector
\begin{equation}
	m_\theta(\mathbf{x}_+) = \begin{bmatrix} m_\theta(x_{+,1}) , \ldots , m_\theta(x_{+,n}) \end{bmatrix}^\top,
\end{equation}
and the covariance matrix
\begin{equation}
	k_\theta(\mathbf{x}_+,\mathbf{x}_+') =
	\begin{bmatrix}
		k_\theta(x_{+,1},x_{+,1}) & \ldots & k_\theta(x_{+,1},x_{+,n}) \\
		\vdots & \ddots & \vdots \\
		k_\theta(x_{+,n},x_{+,1}) & \ldots & k_\theta(x_{+,n},x_{+,n})
	\end{bmatrix}
\end{equation}
leading to the optimized parameters $\bm{\theta}^*$.
Then, the posterior distribution at a location $x$ is given as
\begin{equation} \label{eq:computeMeanPosterior}
\begin{split}
	f(x) \mid \mathbf{x}_+,\mathbf{y}_+,\bm{\theta}^* \sim \mathcal{N}( & m_\theta(x) + k_\theta(x,\mathbf{x}_+) k_\theta(\mathbf{x}_+,\mathbf{x}_+)^{-1} (\mathbf{y}_+ - m_\theta(\mathbf{x}_+)), \\
	& k_\theta(x,x) - k_\theta(x,\mathbf{x}_+) k_\theta(\mathbf{x}_+,\mathbf{x}_+)^{-1} k_\theta(x,\mathbf{x}_+)^\top) 
\end{split}
\end{equation}

This algorithm can be used to predict the damage states of a system. Each time a new currently observed data point is available, the GPM is optimized, and its posterior distribution computed, leading to a new prediction for the upcoming states. However, no prior information based on previous data is considered. Therefore, we present in Section\,\ref{sec:trainGPMprev} how to integrate prior knowledge into a predefined GPM. Furthermore, in this section, the mean and covariance functions have to be chosen a priori involving the assumption of a specific GPM family. As this can lead to poor predictions, we present in Section\,\ref{sec:infer} an approach for inferring an appropriate GPM from previous data and thus weaken the assumption about the GPM family.


\section{Integrating prior knowledge into Gaussian processes}
\noindent In the following subsections, we present two methods for integrating prior knowledge into Gaussian processes. Section\,\ref{sec:trainGPMprev} shows how to train a predefined GPM with previous data, whereas Section\,\ref{sec:infer} proposes an approach for inferring the mean and covariance functions from previous data without using a predefined GPM. In doing so, we assume we have $m$ previously measured or simulated trajectories $\lbrace (x_i,y_i) \mid i=1,...,n_j \rbrace_j$ with $j=1,...,m$ that are finite subsets of realizations generated by one Gaussian process. In order to integrate this previous data into a GPM, we propose a different way of training Gaussian processes for PHM problems: instead of training the GPM to the one currently observed trajectory, we first try to describe all possible trajectories connected to the underlying PHM problem by one Gaussian process. Second, the posterior distribution of the Gaussian process based on current data can be computed, leading to an updated prediction, see Equation \ref{eq:computeMeanPosterior}.

\subsection{Training of a prescribed Gaussian process model with previous data} \label{sec:trainGPMprev}
\noindent Instead of optimizing a prescribed GPM only on currently observed data, we can also train the parameters of a prescribed mean and covariance function with previous data. As in Section\,\ref{sec:GPMcurr}, we assume a GPM chosen a priori which is dependent on a set of parameters $\bm{\theta}$. The observation error can be seen as one parameter of this set. With Bayes' rule, the posterior distribution based on a given trajectory $(\mathbf{x}_j,\mathbf{y}_j)$ is given by
\begin{equation}
	p(\bm{\theta} \mid \mathbf{x}_j,\mathbf{y}_j) = \frac{p(\mathbf{x}_j,\mathbf{y}_j \mid \bm{\theta}) \, p(\bm{\theta})}{\int p(\mathbf{x}_j,\mathbf{y}_j \mid \bm{\theta}) \, p(\bm{\theta}) \, d\bm{\theta}} \propto p(\mathbf{x}_j,\mathbf{y}_j \mid \bm{\theta}) \, p(\bm{\theta}),
\end{equation}
where $p(\bm{\theta})$ and $p(\mathbf{x}_j,\mathbf{y}_j \mid \bm{\theta})$ are the prior and the likelihood distribution, respectively. Assuming an uninformative prior and that all $m$ trajectories are independent and identically distributed, the posterior distribution reads
\begin{equation}
	p(\bm{\theta} \mid \mathbf{x}_1,...,\mathbf{x}_m,\mathbf{y}_1,...,\mathbf{y}_m) = \prod_{j=1}^m \frac{p(\mathbf{x}_j,\mathbf{y}_j \mid \bm{\theta})}{\int p(\mathbf{x}_j,\mathbf{y}_j \mid \bm{\theta}) \, d\bm{\theta}} \propto \prod_{j=1}^m p(\mathbf{x}_j,\mathbf{y}_j \mid \bm{\theta})
\end{equation}
and its log marginal likelihood is
\begin{equation}
	\log p(\bm{\theta} \mid \mathbf{x}_1,...,\mathbf{x}_m,\mathbf{y}_1,...,\mathbf{y}_m) \propto \sum_{j=1}^m \log p(\mathbf{x}_j,\mathbf{y}_j \mid \bm{\theta})
\end{equation}
with
\begin{equation}
	p(\mathbf{x}_j,\mathbf{y}_j \mid \bm{\theta}) = \frac{1}{\sqrt{(2\,\pi)^{n_j}\,\det k_\theta(\mathbf{x}_j,\mathbf{x}_j')}} \exp \left(-\frac{1}{2}\left(\mathbf{y}_j-m_\theta(\mathbf{x}_j)\right)^\top k_\theta(\mathbf{x}_j,\mathbf{x}_j')^{-1}\left(\mathbf{y}_j-m_\theta(\mathbf{x}_j)\right)\right).
\end{equation}
Therefore, the sum of the trajectories' log marginal likelihoods must be maximized, as also stated in \cite{minka1997learning}. In doing so, prior knowledge is integrated into the GPM. However, we still have to choose a predefined mean and covariance function and thus assume a specific GPM family. Therefore, we propose an approach for inferring the mean and covariance functions from previous data in the following. 

\subsection{Inferring Gaussian process models from previous data} \label{sec:infer}
\noindent In contrast to the previous sections, no GPM is chosen a priori. The approach presented in this subsection aims to weaken the assumption about the GPM family by inferring an appropriate GPM from previous data. Instead of implicitly making assumptions about the system output by choosing a mean and a covariance function, we explicitly specify which (basis) functions the system output consists of. In doing so, we assume that each realization $f_j(x)$ can be approximated using a linear combination of a finite set of basis functions $\phi_k(x)$ with $k = 1,...,p$ so that
\begin{equation}
	f_j(x) = \sum_{k=1}^{p} \phi_k(x) \beta_{k,j} = \bm{\phi}(x)^\top \bm{\beta}_j,
\end{equation}
where $\beta_{k,j}$ is the coefficient of realization $j$ belonging to the basis function $\phi_k(x)$. Then, the mean function reads
\begin{align} \label{eq:meanfunction}
	m(x) = \mathbb{E} \left[f(x)\right] = \bm{\phi}(x)^\top \bm{\mu_\beta} \approx \bm{\phi}(x)^\top \bm{\hat{\mu}_\beta}
\end{align}
and the covariance function is
\begin{align} \label{eq:covariancefunction}
	k(x,x') = \mathbb{E} \left[ \left(f(x)-m(x)\right) \left(f(x')-m(x')\right) \right] = \bm{\phi}(x)^\top \bm{\Sigma_\beta} \bm{\phi}(x') \approx \bm{\phi}(x)^\top \bm{\hat{\Sigma}_\beta} \bm{\phi}(x')
\end{align}
where $\bm{\phi}(x)$ is a vector consisting of all basis functions
\begin{equation}
	\bm{\phi}(x) = 
	\begin{bmatrix} \phi_1(x) , \ldots , \phi_p(x) \end{bmatrix}^\top,
\end{equation}
$\bm{\hat{\mu}_\beta}$ is the sample mean vector
\begin{equation}
	\bm{\hat{\mu}_\beta} = \frac{1}{m} \sum_{j=1}^m \bm{\beta}_j = 
	\begin{bmatrix} \hat{\mu}_{\beta_1} , \ldots , \hat{\mu}_{\beta_p} \end{bmatrix}^\top,
\end{equation}
and $\bm{\hat{\Sigma}_\beta}$ is the sample covariance matrix
\begin{equation} \label{eq:sampleCovariance}
	\bm{\hat{\Sigma}_\beta} = 
	\begin{bmatrix}
		\hat{\sigma}_{1,1} & \cdots & \hat{\sigma}_{1,p} \\
		\vdots & \ddots & \vdots \\
		\hat{\sigma}_{p,1} & \cdots & \hat{\sigma}_{p,p}
	\end{bmatrix},
	\hat{\sigma}_{k,l} = \frac{1}{m-1} \sum_{j=1}^m (\beta_{k,j} - \hat{\mu}_{\beta_k}) (\beta_{l,j} - \hat{\mu}_{\beta_l})
\end{equation}
of the coefficients $\bm{B}=[\bm{\beta}_1,...,\bm{\beta}_m]$. The mean function $m(x)$ and the covariance function $k(x,x')$ are derived in the appendix (see Equations\,\ref{eq:meanDerived} and \ref{eq:covDerived}). The approach is similar to assuming a Bayesian linear regression model, the weights' distribution to be multivariate normal (see \cite{Rasmussen.2008}), and approximating the distribution by the sample mean vector and the sample covariance matrix. Furthermore, this idea is similar to modeling the residuals of a global linear model by a GP, which was explored as early as 1975 for polynomials by \cite{blight1975bayesian} and for basis functions by \cite{o1978curve}.

To apply the approach presented, let $\mathbf{x}_j \in \mathbb{R}^{n_j}$ be the previously gathered inputs  and $\mathbf{y}_j \in \mathbb{R}^{n_j}$ the corresponding function values, where trajectory $j$ consists of $n_j$ data points. We first approximate each trajectory $(\mathbf{x}_j,\mathbf{y}_j)$ using a linear regression with $p$ basis functions, where $\bm{\hat{\beta}}_j$ are the estimated coefficients of trajectory $j$ with
\begin{equation} \label{eq:pseudoInverse}
	\bm{\hat{\beta}}_j = (\bm{\Phi}_j^\top \bm{\Phi}_j)^{-1} \bm{\Phi}_j^\top \mathbf{y}_j
\end{equation}
and
\begin{equation}
	\bm{\Phi}_j = \bm{\Phi}(\mathbf{x}_j) = 
	\begin{bmatrix}
	\phi_1(x_{1,j}) & \cdots & \phi_p(x_{1,j}) \\	
	\vdots & \ddots & \vdots \\
	\phi_1(x_{n_j,j}) & \cdots & \phi_p(x_{n_j,j})
	\end{bmatrix},
\end{equation}
where $n_j \geq p$ and $m > p$. If $n_j < p$, $(\bm{\Phi}_j^\top \bm{\Phi}_j)$ becomes rank deficient and therefore has no inverse, which causes problems in Equation\,\ref{eq:pseudoInverse}. One could use the Moore-Penrose pseudoinverse instead. If $m \leq p$, $\mathrm{rank}(\bm{\hat{\Sigma}_a})=m-1$ and therefore the sample covariance matrix has no inverse. This problem could be solved by adding a small perturbation to the main diagonal of the sample covariance matrix so that the sample covariance matrix has full rank.

Now we are able to compute the sample mean vector $\bm{\hat{\mu}_{\hat{\beta}}}$ and the sample covariance matrix $\bm{\hat{\Sigma}_{\hat{\beta}}}$ of the estimated coefficients $\bm{\hat{B}}$ leading to an approximate mean (Equation\,\ref{eq:meanfunction}) and covariance function (Equation\,\ref{eq:covariancefunction}). These functions are inferred from previous data and thus incorporate prior knowledge and weaken the assumption about the GPM family.

Additionally, the observation error $\sigma_y$ can be estimated using the root mean square of the residuals $\hat{r}$
\begin{align}
	\mathbf{\hat{r}}_j = & \bm{\Phi}_j \bm{\hat{\beta}}_j - \mathbf{y}_j \\
	\hat{\sigma}_y = & \sqrt{\frac{1}{m} \sum_{j=1}^m \frac{1}{n_j} \sum_{i=1}^{n_j} \hat{r}_{i,j}^2} \label{eq:observationError}
\end{align}
or by maximizing the sum of the trajectories' log marginal likelihoods (Equation\,\ref{eq:maximizeLogMargLike}) with the covariance matrix $\mathbf{K}=k(\mathbf{x}_j,\mathbf{x}_j') + \sigma_y^2 \, \mathbf{I}$. In the following, we apply all three methods proposed, i.e., Gaussian processes with (1) current data \& a prescribed GPM, (2) previous data \& a prescribed GPM, and (3) previous data \& basis functions, to four PHM data sets in order to show the impact of the different assumptions.


\section{Effect of integrating prior knowledge} \label{sec:poly}
\begin{figure}[h]
\centering
	\begin{tabular}{c c}
	\includegraphics*[scale=1.0]{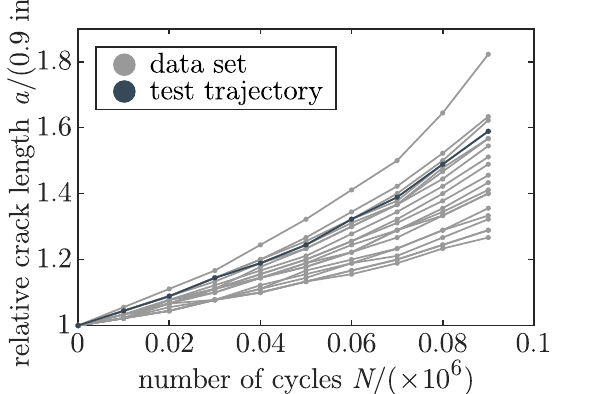} & \includegraphics*[scale=1.0]{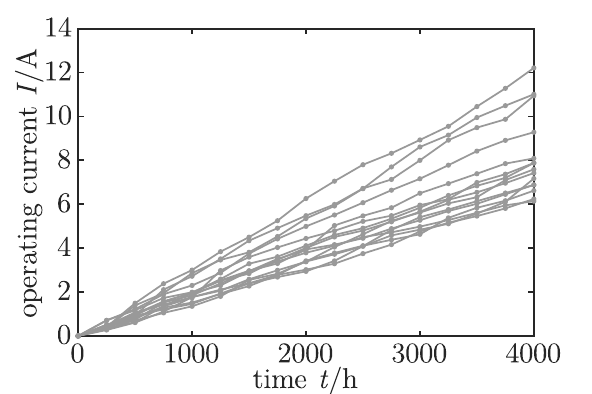} \\
	(a) & (b) \\
	\includegraphics*[scale=1.0]{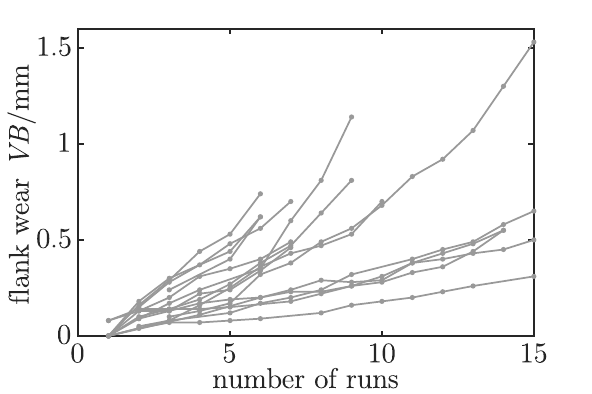} & \includegraphics*[scale=1.0]{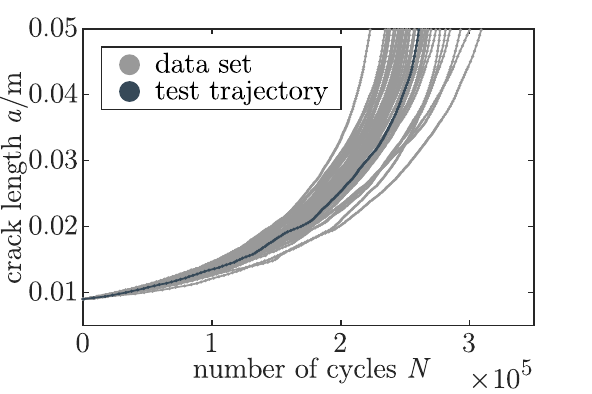} \\
	(c) & (d) \\
	\end{tabular}
  \caption{PHM data sets representing (a) fatigue crack growth, (b) laser degradation, (c) milling machine wear, and (d) fatigue crack growth (Virkler).}
  \label{fig:dataSets}
\end{figure}

\noindent In this section, we investigate the effect of integrating prior knowledge into Gaussian processes. In order to compare the three methods proposed, we apply them to published PHM data sets, which represent fatigue crack growth, laser degradation, and milling machine wear. The three data sets are shown in Figure\,\ref{fig:dataSets}. A subset of the fatigue crack growth data set published by \citet{virkler1979statistical}, see Figure\,\ref{fig:dataSets}d, is investigated in Section\,\ref{sec:physics}. The two highlighted test trajectories in Figures\,\ref{fig:dataSets}a and \ref{fig:dataSets}d are used to visualize the results in the following sections.

The fatigue crack growth data used comes from \citet{hudak1978development} and was visually obtained from figure 4.5.2 on page 242 of \citet{bogdanoff1985probabilistic} by \citet{lu1993using}. The data set in Figure\,\ref{fig:dataSets}a, consists of 21 fatigue crack growth trajectories where the crack length $a$ is given every $10,000$ cycles and cut at $90,000$ cycles. The crack lengths are normalized by the initial crack length of $a_0=0.9\,\mathrm{in}$.

The laser degradation data set was published by \citet{meeker2014statistical} and consists of 15 GaAs (gallium arsenide) laser device degradation trajectories. In general, the output light intensity will decrease over time. However, due to a feedback mechanism, the laser output is kept constant by an increasing operating current. The operating current $I$ was measured every $250\,\mathrm{h}$ until $4000\,\mathrm{h}$. The trajectories are shown in Figure\,\ref{fig:dataSets}b.

The milling machine wear data set was published by \citet{agogino2007milling}. The data set consists of several sensor measurements. In this study, however, we only use the flank wear over the executed runs. In total, 16 cases were investigated. Since one case comprises only an initial value, we use 15 milling machine wear trajectories for our study, see Figure\,\ref{fig:dataSets}c.

For comparing the methods proposed, underlying models have to be assumed. As the data sets of Figures\,\ref{fig:dataSets}a-c follow a polynomial behavior predominantly, we assume polynomial mean and covariance functions in the following. Additionally, a zero mean and a squared-exponential covariance function are considered as they have been frequently used. In order to find the maximum polynomial order, we split each previous data trajectory into a train ($70\,\%$) and a test set ($30\,\%$) and evaluate the mean squared error for the test set. The smallest test set error is obtained with a maximum order of $q=1$ for the laser degradation and the milling machine wear data set, and an order of $q=2$ for the fatigue crack growth data set. Table\,\ref{tb:models} summarizes the models explained in the following.

\begin{table}[h]
\caption{Investigated models.}
\centering
\resizebox{\textwidth}{!}{%
\begin{tabular}{l l l l l l}
	\hline
	\textbf{Abbreviation} & \textbf{Model} & \textbf{Mean function} & \textbf{Covariance function} & \textbf{\multirow{2}{3.5cm}{Data for parameter \\ estimation}} & \textbf{\multirow{2}{3.5cm}{Data for posterior \\ distribution}} \\ \\
	\hline
	GPM curr.          & predefined GPM  & zero       & polynomial          & current data  & current data \\
	GPM prev. ZM \& SE & predefined GPM  & zero       & squared-exponential & previous data & current data \\
	GPM prev. POLY     & predefined GPM  & polynomial & polynomial          & previous data & current data \\
	IGPM               & basis functions & polynomial & polynomial          & previous data & current data \\
	
	\hline
\end{tabular}} \label{tb:models}
\end{table}

For \textit{training GPMs with current data} (GPM curr.), we use only the currently observed data points and no previous trajectories, i.e., each trajectory is predicted without knowing the other ones. For this method, we assume a zero mean and a polynomial covariance function with the above determined maximum polynomial order\,$q$
\begin{equation} \label{eq:polyCov}
	k_\theta(x,x')=\sigma_f^2 (x x' + b)^q + \sigma_y^2 \mathbf{I},
\end{equation}
where $\sigma_f$, $b$, and $\sigma_y$ are the parameters $\bm{\theta}$ of the GPM. These parameters $\bm{\theta}$ are trained for every new data point by maximizing the log marginal likelihood. We found that using a polynomial mean function in this particular setup decreases the accuracy, which might be a result of too many parameters for a relatively small quantity of data. Moreover, the optimization starting point strongly influences the result. Therefore, it is useful to start the optimization with a set of different starting points. To be able to compare the results to the other methods, we use the parameters optimized to the previous data as the starting point for this method.

For \textit{training a GPM with previous data} (GPM prev.), we assume all trajectories but the currently investigated one to be previously observed. Due to the larger amount of data, the assumed GPM consists of a polynomial covariance function, see Equation\,\ref{eq:polyCov}, and a polynomial mean function
\begin{equation}
	m_\theta(x)=\sum_{k=1}^{q+1} c_k x^{k-1},
\end{equation}
where $\sigma_f$, $b$, and $c_k$ are the parameters $\bm{\theta}$ of the GPM (GPM prev. POLY). Additionally, a GPM with a zero mean and a squared-exponential function, see Equation\,\ref{eq:se}, is trained with previous data (GPM prev. ZM \& SE) in order to investigate the influence of the prescribed GPM. As explained in Section\,\ref{sec:trainGPMprev}, we maximize the sum of the trajectories' log marginal likelihood for optimizing the parameters $\bm{\theta}$. These parameters are fixed for the entire prediction series. For this method, the currently observed data acts as a conditional on the Gaussian process leading to an updated prediction.

For \textit{inferring the GPM} (IGPM), we again assume all trajectories but the currently investigated one to be previously observed. In order to determine the GPM, first, each previous trajectory is approximated by linear regression of polynomial basis functions with the maximum order $q$, see Equation\,\ref{eq:pseudoInverse}. Second, the estimated coefficients are used to determine the mean and the covariance function, see Equations\,\ref{eq:meanfunction}-\ref{eq:sampleCovariance}. Additionally, the observation error is estimated with Equation\,\ref{eq:observationError}. As for the GPM trained with previous data, the resulting GPM and the observation error are fixed for the entire prediction series. After observing a current data point, the posterior distribution of the Gaussian process is computed, leading to an updated prediction, see Equation\,\ref{eq:computeMeanPosterior}.

\subsection{Results} \label{sec:polyResults}
\begin{figure}[]
\centering
	\begin{tabular}{c c}
	\includegraphics*[scale=1.0]{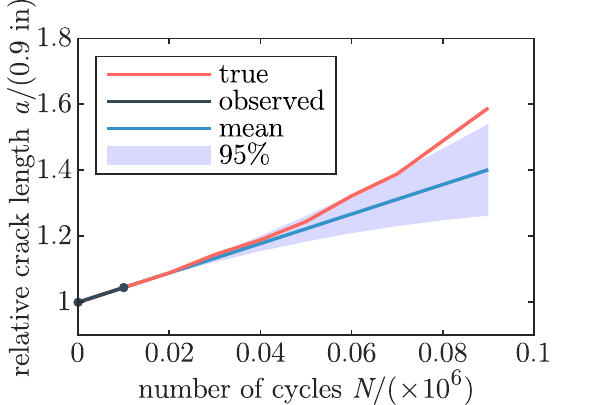} & \includegraphics*[scale=1.0]{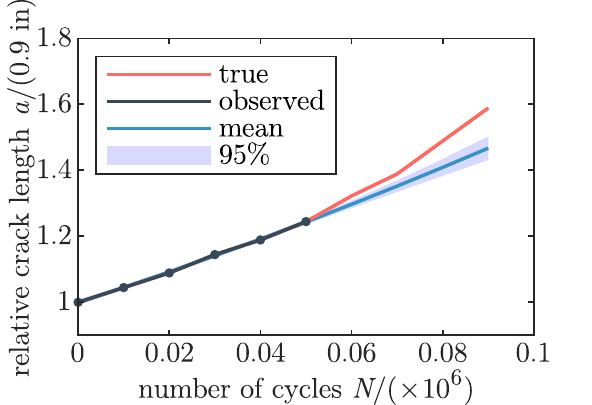} \\
	(a) & (b) \\
	\includegraphics*[scale=1.0]{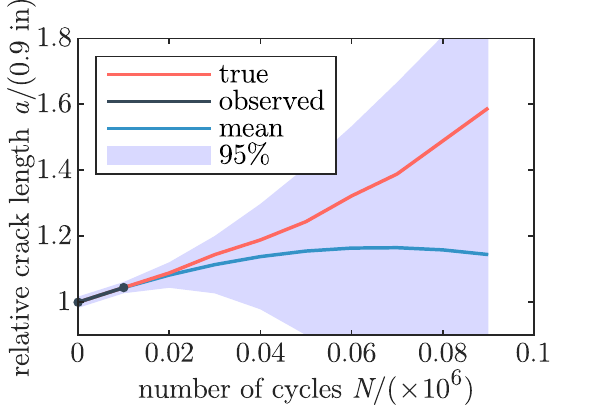} & \includegraphics*[scale=1.0]{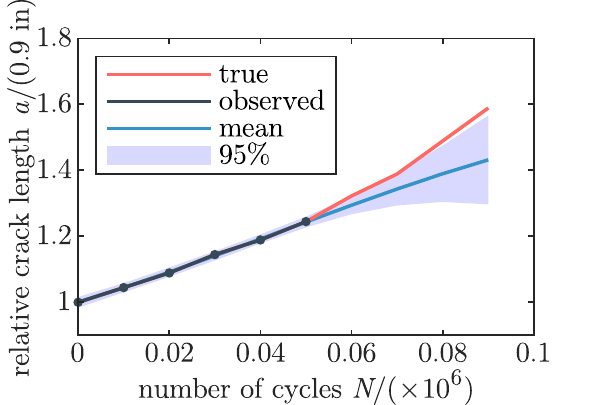} \\
	(c) & (d) \\
	\includegraphics*[scale=1.0]{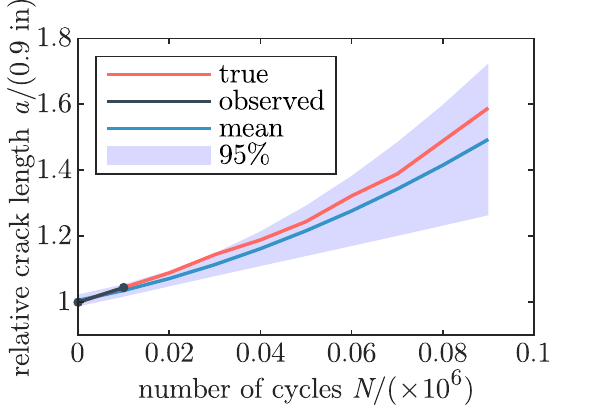} & \includegraphics*[scale=1.0]{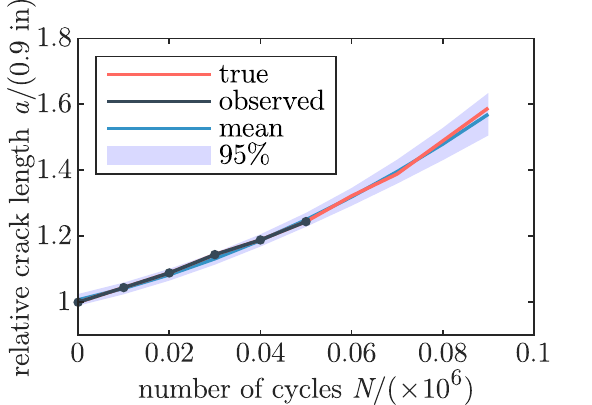} \\
	(e) & (f) \\
	\includegraphics*[scale=1.0]{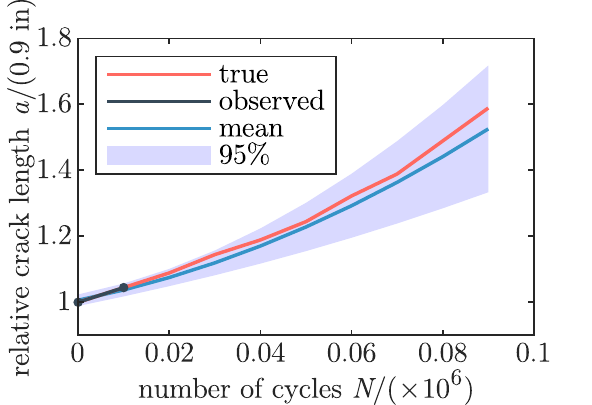} & \includegraphics*[scale=1.0]{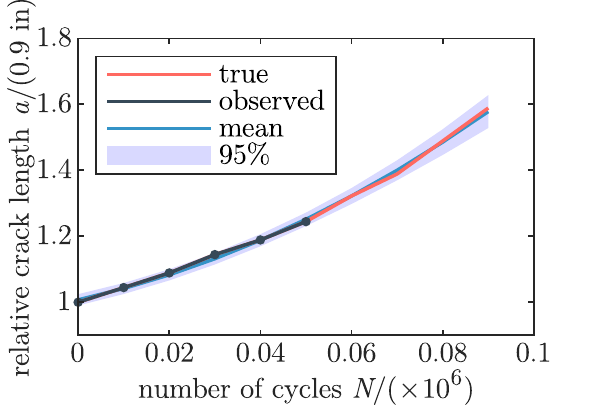} \\
	(g) & (h) \\
	\end{tabular}
  \caption{Probabilistic predictions of crack growth at two different time states. (a) GPM curr.: GPM trained with current data for an early and (b) a late time state. (c) GPM prev. ZM \& SE: Zero mean and squared-exponential function GPM trained with previous data for an early and (d) a late time state. (e) GPM prev. POLY: Polynomial GPM trained with previous data for an early and (f) a late time state. (g) IGPM: Inferred GPM for an early and (h) a late time state.}
  \label{fig:CompareFCGPred}
\end{figure}

\noindent The probabilistic predictions for the fatigue crack growth data are shown in Figure\,\ref{fig:CompareFCGPred}. The figure visualizes the true trajectory (red), the mean prediction (blue), and the symmetric 95\% credible region (light blue area), i.e., the region between the 2.5\% and the 97.5\% quantiles. Figure\,\ref{fig:CompareFCGPred} compares the prediction of the GPM, which relies only on current data, with the GPMs trained with previous data (GPM prev. ZM \& SE and GPM prev. POLY), and with the inferred GPM. Two different time states of the currently observed trajectory are visualized.


\begin{table}[h]
\caption{Errors for different methods with respect to the corresponding data set.}
\centering
\resizebox{\textwidth}{!}{%
\begin{tabular}{c l l l r l l l r l r l r l}
	\hline
	 \textbf{Data Set} & \textbf{Model} & \multicolumn{2}{c}{\textbf{RMSE}} & \multicolumn{2}{c}{\textbf{MAPE}} & \multicolumn{2}{c}{\textbf{RMSE half}} & \multicolumn{2}{c}{\textbf{MAPE half}} & \multicolumn{2}{c}{\textbf{Prediction time}} & \multicolumn{2}{c}{\textbf{Selection time}} \\
	 
	 \hline
	 
	 \multirow{4}{*}{\thead{Fatigue Crack \\ Growth}} & IGPM & 0.06 & & 0.03 & & 0.02 & & 0.01 & & $0.003\,\mathrm{s}$ & & $0.16\,\mathrm{s}$ & \\
	 & GPM curr. & 0.19 & $+216.8\%$ & 0.09 & $+207.2\%$ & 0.06 & $+175.3\%$ & 0.03 & $+162.1\%$ & $0.91\,\mathrm{s}$ & $+26,797.1\%$ & $0.00\,\mathrm{s}$ & $-100.0\%$ \\
	 & GPM prev. ZM \& SE & 0.29 & $  +395.6\%$ & 0.13 & $ +383.8\%$ & 0.08 & $  +270.7\%$ & 0.04 & $  +237.1\%$ & $0.01\,\mathrm{s}$ & $+261.8\%$ & $1.46\,\mathrm{s}$ & $+786.7\%$ \\
	 & GPM prev. POLY & 0.06 & $  +9.4\%$ & 0.03 & $ +15.9\%$ & 0.02 & $  +9.8\%$ & 0.01 & $  +9.1\%$ & $0.02\,\mathrm{s}$ & $+352.9\%$ & $1.78\,\mathrm{s}$ & $+982.8\%$ \\
	
	\hline
	
	\multirow{4}{*}{\thead{Laser \\ Degradation}} & IGPM & $0.90\,\mathrm{A}$ & & 0.09 & & $0.42\,\mathrm{A}$ & & 0.05 & & $0.004\,\mathrm{s}$ & & $0.18\,\mathrm{s}$ & \\
	 & GPM curr. & $2.28\,\mathrm{A}$ & $+151.9\%$ & 0.15 & $ +68.2\%$ & $0.41\,\mathrm{A}$ & $  -2.1\%$ & 0.05 & $  -1.2\%$ & $2.45\,\mathrm{s}$ & $+59,624.4\%$ & $0.00\,\mathrm{s}$ & $-100.0\%$ \\
	& GPM prev. ZM \& SE & $2.44\,\mathrm{A}$ & $  +170.1\%$ & 0.17 & $  +96.9\%$ & $0.58\,\mathrm{A}$ & $  +38.5\%$ & 0.06 & $  +20.4\%$ & $0.02\,\mathrm{s}$ & $+373.2\%$ & $1.00\,\mathrm{s}$ & $+465.1\%$ \\	
	 & GPM prev. POLY & $0.91\,\mathrm{A}$ & $  +0.3\%$ & 0.09 & $  +0.1\%$ & $0.41\,\mathrm{A}$ & $  -1.4\%$ & 0.05 & $  -1.0\%$ & $0.02\,\mathrm{s}$ & $+490.2\%$ & $2.72\,\mathrm{s}$ & $+1,441.1\%$ \\
	
	\hline
	
	\multirow{4}{*}{\thead{Milling \\ Machine Wear}} & IGPM & $0.23\,\mathrm{mm}$ & & 0.27 & & $0.14\,\mathrm{mm}$ & & 0.17 & & $0.0009\,\mathrm{s}$ & & $0.17\,\mathrm{s}$ & \\
	 & GPM curr. & $0.31\,\mathrm{mm}$ & $+39.8\%$ & 0.35 & $+28.8\%$ & $0.15\,\mathrm{mm}$ & $ +9.2\%$ & 0.18 & $+11.7\%$ & $4.07\,\mathrm{s}$ & $+465,892.9\%$ & $0.00\,\mathrm{s}$ & $-100.0\%$ \\
	 & GPM prev. ZM \& SE & $0.36\,\mathrm{mm}$ & $ +60.9\%$ & 0.41 & $ +53.6\%$ & $0.17\,\mathrm{mm}$ & $ +25.5\%$ & 0.20 & $ +20.9\%$ & $0.01\,\mathrm{s}$ & $+1,422.4\%$ & $1.04\,\mathrm{s}$ & $+496.0\%$ \\
	 & GPM prev. POLY & $0.23\,\mathrm{mm}$ & $ +1.1\%$ & 0.27 & $ +1.9\%$ & $0.13\,\mathrm{mm}$ & $ -2.9\%$ & 0.16 & $ -5.0\%$ & $0.02\,\mathrm{s}$ & $+1,708.5\%$ & $1.57\,\mathrm{s}$ & $+796.5\%$ \\
	
	\hline

\end{tabular}} \label{tb:polyResults}
\end{table}

Common to all methods is that we obtain an updated prediction after observing a new current data point leading to $n$ predictions for each trajectory. The models are evaluated in a leave-one-out validation scheme, i.e., each trajectory is used as test data while the other ones are used for training. In this study, we evaluate the root mean squared error (RMSE) and the mean absolute percentage error (MAPE) of the last predicted point for one trajectory from prediction $1$ to prediction $n-1$ and compute the average of all investigated trajectories. In order to get an idea of how the accuracy changes with an increasing number of current data points, we also evaluate the RMSE and MAPE starting from prediction $\ceil*{n/2}$. These are referred to as RMSE \textit{half} and MAPE \textit{half}. The evaluated errors for the three data sets are presented in Table\,\ref{tb:polyResults}.
Moreover, the average prediction time for one series, i.e., the total time for all $n-1$ predictions of one trajectory, as well as the selection time for the GPM, i.e., the time for determining/optimizing the used GPM, are listed in Table\,\ref{tb:polyResults}.

\begin{figure}[]
\centering
	\begin{tabular}{c c}
  	\includegraphics*[]{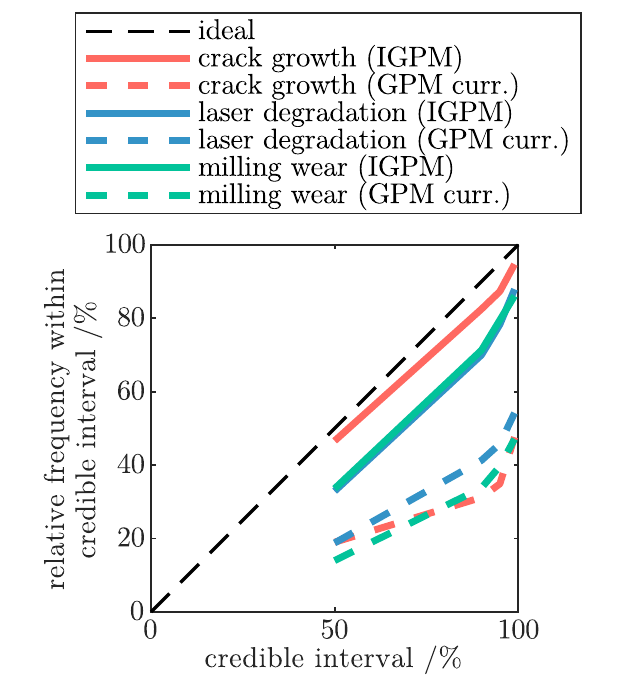} & \includegraphics*[]{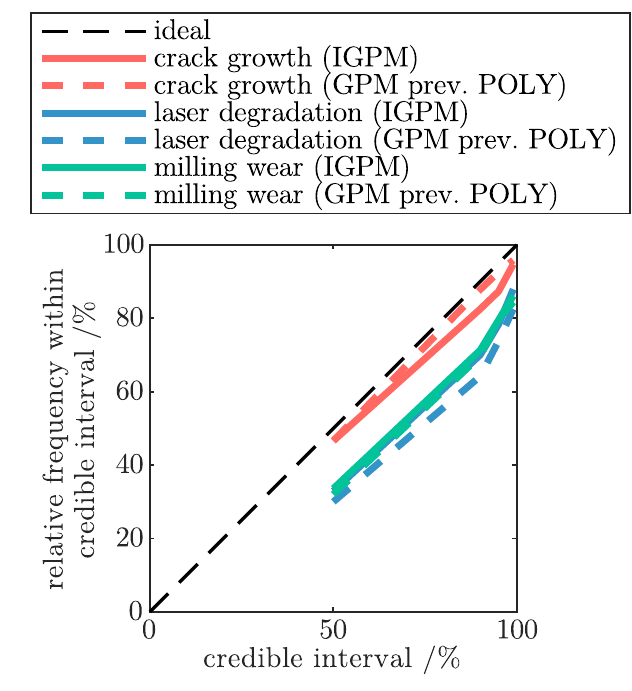} \\
  	(a) & (b)
  	\end{tabular}
\caption{Credible interval comparison of IGPM with (a) GPM trained with current data (GPM curr.), and with (b) polynomial GPM trained with previous data (GPM prev. POLY).}
\label{fig:IGPM-GPcurr}
\end{figure}

Additionally, we evaluated how often the last measurement value lies within a certain predicted credible interval. This opens up the possibility of quantifying the predicted uncertainty since the relative frequency of the real value lying within the credible interval should correspond to the assumed credible interval itself. The symmetric 95\% credible interval, for example, is the interval between the 2.5\% and the 97.5\% quantile and should correspond to a relative frequency of 95\%. The results are plotted in Figures\,\ref{fig:IGPM-GPcurr}a and \ref{fig:IGPM-GPcurr}b for four symmetric credible intervals $\{50\%,90\%,95\%,99\%\}$ where the inferred GPM is compared to the model trained with current data and to the polynomial GPM trained with previous data, respectively. A value above the black dashed line indicates an overly wide credible interval, which can be viewed as a too-conservative prediction. In contrast, a value below the dashed line results from a too-small credible interval, which can be viewed as an overly optimistic prediction.

\subsection{Discussion} \label{sec:polyDiscuss}
\noindent By comparing the IGPM to the GPM trained with current data, we see that the IGPM results in higher accuracy, in particular at the beginning (see Table\,\ref{tb:polyResults}). This is a result of using previous data since the IGPM gains information about the shape of typical trajectories. This information is embedded in the mean function. Furthermore, the information about how different locations are correlated and distributed is embedded in the covariance function. The error margin decreases towards the end of a prediction series (see RMSE half and MAPE half) because of the larger amount of current data, which enables better parameter training. At the beginning of the prediction series, the ratio of the number of parameters to current data is high, and the optimization problem is prone to overfit, whereas the ratio decreases towards the end. Moreover, Figures\,\ref{fig:CompareFCGPred}a and \ref{fig:CompareFCGPred}b show that training a GPM only with current data leads to a shorter look-ahead time, whereas the IGPM is able to accurately predict even points that are far away, see Figures\,\ref{fig:CompareFCGPred}g and \ref{fig:CompareFCGPred}h. On the other hand, for the laser degradation data set, the RMSE half and MAPE half of the IGPM are even larger than the ones based on the GPM trained with current data (see Table\,\ref{tb:polyResults}). This might be a result of the estimated observation error, which is flexible within a prediction series of one trajectory for the GPM trained with current data. Another reason for this behavior is that the laser degradation set represents a relatively simple degradation behavior that is predominantly linear. For more complex data sets, e.g., fatigue crack growth and milling machine wear, the RMSE half and MAPE half are larger.

Furthermore, the GPM trained with current data tends to underestimate the uncertainty for PHM problems. As Figure\,\ref{fig:IGPM-GPcurr}a shows, the colored dashed lines (GPM curr.) are further away from the black dashed line than the solid lines (IGPM), resulting from a variance that was underestimated, a wrong mean, or both. Again, this is due to the lack of previous data. There is no prior information about how large the scatter is, resulting in wrong assumptions/estimations of the GPM's parameters. In contrast, this information is integrated into the inferred covariance function of the IGPM.

By comparing the prediction times, we see that the IGPM obtains results more quickly. The reason for this is that only the posterior distribution is computed for new current observations. By contrast, for the GPM trained with current data, an optimization problem has to be additionally solved every time a new current data point is observed. The selection time for the IGPM is, of course, larger due to the fact that for the GPM trained with current data, no previous selection process is carried out. 

By comparing the IGPM to the polynomial GPM trained with previous data, we see that their results are very similar. This is due to the fact that their determined mean functions are identical, and their covariance functions are almost the same. The IGPM tends to have slightly higher accuracy, see Figures\,\ref{fig:CompareFCGPred}e-h and Table\,\ref{tb:polyResults}, because the model captures the correlation between the coefficients. This improves accuracy if, for example, the coefficient of order 1 is correlated with the coefficient of order 2. By writing the covariance functions for $q=2$ explicitly, the difference can be identified:
\begin{align*}
	k_{prev}(x,x') =& \sigma_f^2 \, b^2 + &2 \, \sigma_f^2 \, b \, x x'& + &\sigma_f^2 \, (x x')^2 \\
	k_{IGPM}(x,x') =& (\mathbb{E}[\beta_0^2] - \mathbb{E}[\beta_0]^2) + &(\mathbb{E}[\beta_1^2] - \mathbb{E}[\beta_1]^2) \, x x'& + &(\mathbb{E}[\beta_2^2] - \mathbb{E}[\beta_2]^2) \, (x x')^2 \\
	& + (\mathbb{E}[\beta_1 \, \beta_0] - \mathbb{E}[\beta_1] \, \mathbb{E}[\beta_0]) \, (x + x') \\
	& + (\mathbb{E}[\beta_2 \, \beta_0] - \mathbb{E}[\beta_2] \, \mathbb{E}[\beta_0]) \, (x^2 + x'^2) \\
	& + (\mathbb{E}[\beta_2 \, \beta_1] - \mathbb{E}[\beta_2] \, \mathbb{E}[\beta_1]) \, (x^2 x' + x'^2 x)
\end{align*}
Of course, parameters could be added to the modeled covariance function in order to capture the correlation. This, however, would lead to a more complex optimization problem due to the possibility of including further local optima. Ultimately, the predicted mean and variance are also almost the same, see Figure\,\ref{fig:IGPM-GPcurr}b, as the mean and the covariance functions are similar.

Additionally, a prescribed GPM with a zero mean and a squared-exponential function is trained with previous data. Due to the poorly chosen GPM, the accuracy is significantly worse than the accuracy of the IGPM method. As the mean function is set to zero, the mean prediction converges to zero for points that are far away from the currently measured data, see Figure\,\ref{fig:CompareFCGPred}c. For close points, the error is smaller, see Table\,\ref{tb:polyResults} and Figure\,\ref{fig:CompareFCGPred}d. Ultimately, it can be seen that the prescribed GPM chosen has a significant influence on the accuracy.

By looking at the selection time, we see that the IGPM method is faster than training a predefined GPM. This is due to the fact that, in general, optimizing the parameters of a GPM has a non-convex objective function \cite{Rasmussen.2008}. In contrast, for the IGPM method, the coefficients of a linear combination of basis functions can be estimated by solving a linear system of equations. Furthermore, we found that the starting point of the GPM optimization problem significantly influences the results. For a couple of runs, the optimized parameters led to a completely different covariance function which resulted in poor predictions. Since the objective function is, in general, non-convex, a gradient-based optimizer can lead to local optima.

\begin{figure}[h]
	\centering
	\includegraphics*[scale=1.0]{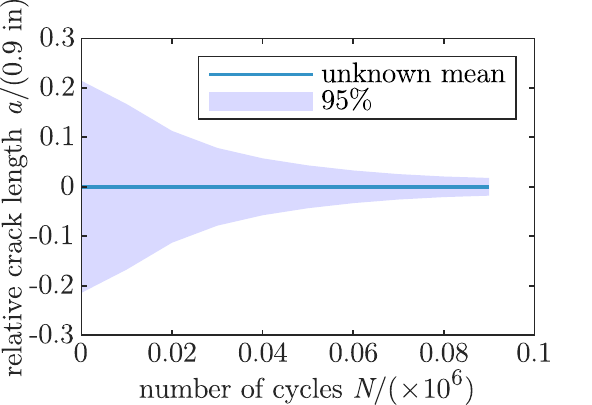}
	\caption{Predicted symmetric 95\% credible interval of the crack length at $90,000$ cycles. The x-axis indicates at which time the prediction is made. The shown credible interval is known before any current data point is observed. Yet, the predicted mean of the crack length is unknown as it depends on the observed function values.}
  	\label{fig:predCiFatigueCrackGrowth}
\end{figure}

Furthermore, we recognized an interesting property of GPMs with fixed parameters, i.e., all free parameters including the observation error are determined: the variance of the posterior distribution is known over the entire range of $x$ once the GPM, the observation error, and the measurement locations are determined. This is due to the fact that the (co-)variance of a conditioned Gaussian distribution is independent of the conditioning values, see Equation\,\ref{eq:computeMeanPosterior}. Therefore, we know for every point of time with what certainty we will be able to predict the system output before we have observed any current data point. This is not possible if the GPM is re-optimized for every new current data point because this changes the GPM's parameters. Figure\,\ref{fig:predCiFatigueCrackGrowth} shows the symmetric 95\% credible interval of the crack length at $90,000$ cycles predicted at different times. The x-axis shows the point in time up to which current data is measured. The credible interval at $90,000$ cycles shrinks as more and more data is collected. This figure can be created before any data point is measured as the variance of the posterior distribution depends only on the measurement locations, i.e., the number of cycles at which new data is observed. Yet, the mean depends also on the measured function values, which is why it is not known beforehand.


\section{Effect of using physics-informed basis functions} \label{sec:physics}
\noindent For the IGPM method, see Equations\,\ref{eq:meanfunction} and \ref{eq:covariancefunction}, we can also choose other types of basis functions. In this section, we apply physics-informed basis functions to a set of published fatigue crack growth data and examine the effect of using such basis functions. The measurements of the investigated data set were performed by Virkler in 1977 \cite{virkler1979statistical} and consist of 68 crack growth trajectories, see Figure\,\ref{fig:dataSets}d. The center-cracked aluminum 2024-T3 test specimens with a total width of $W=152.4 \, \mathrm{mm}$ and an initial crack size of $a_0=9 \, \mathrm{mm}$ were tested under a constant amplitude loading with a stress range of $\Delta \sigma_\infty = 48.26 \, \mathrm{MPa}$ up to a final crack length of $a_f=49.8 \, \mathrm{mm}$.  During the test, the number of cycles was measured every $\Delta a=0.2 \, \mathrm{mm}$ up to $a=36.2 \, \mathrm{mm}$, every $\Delta a=0.4 \, \mathrm{mm}$ up to $a=44.2 \, \mathrm{mm}$, and every $\Delta a=0.8 \, \mathrm{mm}$ up to $a=49.8 \, \mathrm{mm}$.

The crack growth rate can be modeled by Paris' law \cite{paris1963critical}
\begin{equation}
	\frac{da}{dN} = C (\Delta K)^\alpha,
\end{equation}
where $C$ and $\alpha$ are material properties and $\Delta K$ is the range of the stress intensity factor. According to \cite{astm1981test}, for a center-cracked metal sheet with a finite width $W$, $\Delta K$ can be computed as
\begin{equation}
	\Delta K = \frac{\Delta \sigma_\infty \sqrt{\pi a}}{\sqrt{\mathrm{cos}\left(\frac{\pi a}{W}\right)}}.
\end{equation}
Assuming that $C$ and $\alpha$ are independent of $a$, with the initial condition $N_0=0$, the differential equation can be solved for the number of cycles $N$ as
\begin{equation}
	N(a) = \frac{1}{C \, \Delta \sigma_\infty^\alpha \, \pi^{\alpha/2}} \int_{a_0}^{a} \left(\frac{\mathrm{cos}\left(\frac{\pi \, z}{W}\right)}{z} \right)^{\alpha/2} dz.
\end{equation}
Therefore, we use $\phi \equiv N$ as our basis function. According to \cite{wang2017fatigue}, the exponent is set to $\alpha=2.9$. The parameter $C$ is not important since we compute a multiplicative coefficient for each trajectory, see Equation\,\ref{eq:pseudoInverse}. However, in order to avoid small values for the coefficients, we set the material parameter to $C=8.7096 \times 10^{-11}$. The trajectories 1-47 are used to infer the GPM. The factor $\sigma_x$ of the observation error
\begin{equation}
	\sigma_y(x) = \sigma_x \abs*{\frac{\mathrm{d}m(x)}{\mathrm{d}x}}
\end{equation}
is optimized by maximizing the log marginal likelihood of the last measured point.

In general, the exponent $\alpha$ might not be the same for all trajectories. To take this into account, we can set several values for $\alpha$, which results in multiple basis functions
\begin{equation}
	\bm{\phi}(x) =  
	\begin{bmatrix} \phi_1(x,\alpha=\alpha_1), \ldots, \phi_p(x,\alpha=\alpha_p) \end{bmatrix}^\top.
\end{equation}
In this study, we apply the method with $\alpha=2.9$ and compare it to a polynomial basis function of order $q=4$ and to the physics-informed basis functions with $\alpha=\{2.6,2.8,3.0,3.2\}$. Actually, from a mathematical point of perspective, it would be necessary to prove linear independence such that the corresponding regression system matrix $\bm{\Phi}_j^\top \bm{\Phi}_j$ is invertible, see Equation\,\ref{eq:pseudoInverse}.

\subsection{Results} \label{sec:physicsResults}
\noindent As in Section\,\ref{sec:poly}, we evaluate the RMSE and the MAPE of the last predicted point. The results are listed in Table\,\ref{tb:physicResults}.

\begin{table}[h]
\caption{Errors for different basis functions with respect to Virkler's crack growth data set.}
\centering
\resizebox{\textwidth}{!}{%
\begin{tabular}{c r l r l r l r l}
	\hline
	\textbf{Model} & \multicolumn{2}{c}{\textbf{RMSE}} & \multicolumn{2}{c}{\textbf{MAPE}} & \multicolumn{2}{c}{\textbf{RMSE half}} & \multicolumn{2}{c}{\textbf{MAPE half}} \\
	 \hline
	 \thead{Physics-informed $\alpha=2.9$} & 9368.90 & & 0.03 & & 3528.50 & & 0.01 & \\
	 \thead{Polynomial $p=4$} & 9059.10 & $-3.3\%$ & 0.03 & $-2.1\%$ & 5157.40 & $+46.2\%$ & 0.02 & $+50.4\%$ \\
	 \thead{Physics-informed $\alpha=\{2.6,2.8,3.0,3.2\}$} & 7376.90 & $-21.3\%$ & 0.02 & $-28.3\%$ & 3266.80 & $ -7.4\%$ & 0.01 & $-10.9\%$ \\
	\hline
\end{tabular}} \label{tb:physicResults}
\end{table}

\begin{figure}[h!]
\centering
	\begin{tabular}{c c}
	\includegraphics*[scale=1.0]{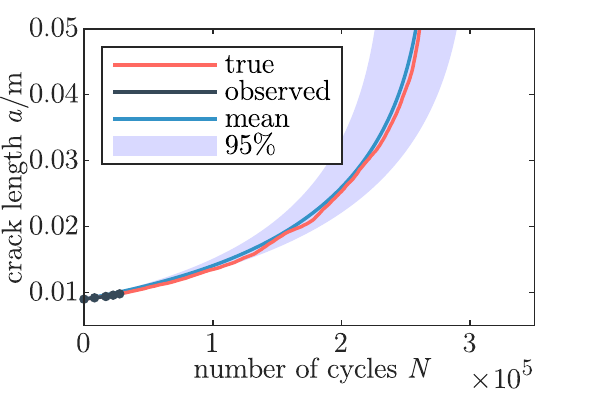} & \includegraphics*[scale=1.0]{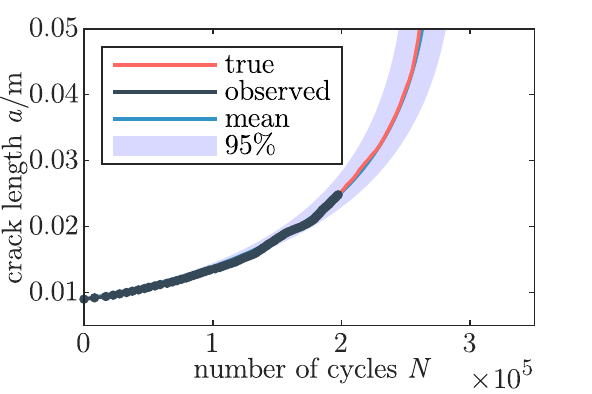} \\
	(a) & (b) \\
	\includegraphics*[scale=1.0]{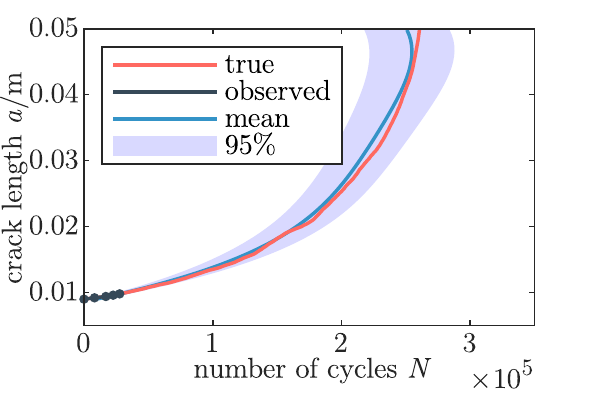} & \includegraphics*[scale=1.0]{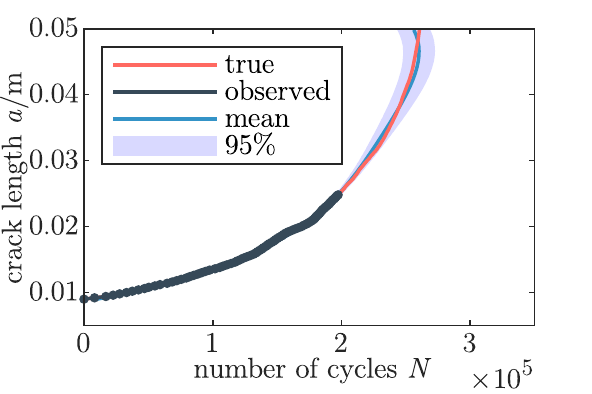} \\
	(c) & (d) \\
	\includegraphics*[scale=1.0]{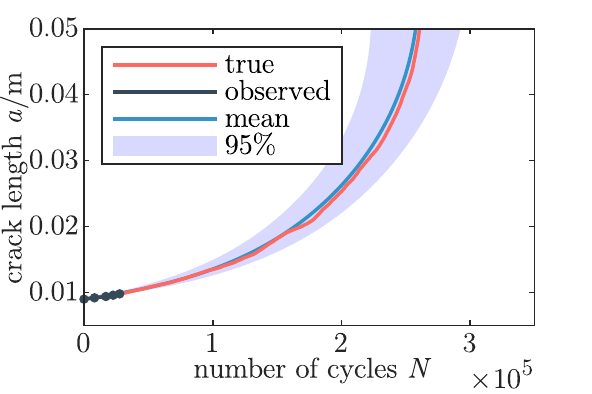} & \includegraphics*[scale=1.0]{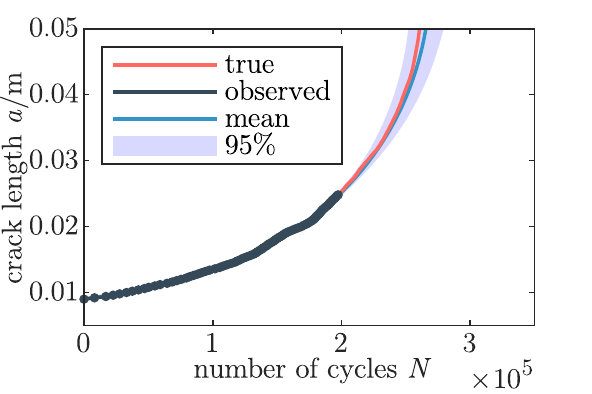} \\
	(e) & (f) \\
	\end{tabular}
  \caption{Probabilistic predictions of crack growth with different basis functions at two different time states. Note that $x=a$ and $f \equiv N$ are flipped. (a) Physics-informed mean and coveriance function with $\alpha=2.9$ for an early and (b) a late time state. (c) Polynomial mean and covariance function with $q=4$ for an early and (d) a late time state. (e) Physics-informed mean and coveriance function with $\alpha=\{2.6,2.8,3.0,3.2\}$ for an early and (f) a late time state.}
  \label{fig:CompareVirkPred}
\end{figure}

\noindent Additionally, the probabilistic predictions for the three different basis functions are shown in Figure\,\ref{fig:CompareVirkPred}. The figure depicts the true trajectory (red), the mean prediction (blue), and the symmetric 95\% credible region (light blue area), i.e., the region between the 2.5\% and the 97.5\% quantiles. Two different time states of the currently observed trajectory are shown.

\subsection{Discussion}
\begin{figure}[h!]
\centering
	\begin{tabular}{c c}
	\includegraphics*[scale=1.0]{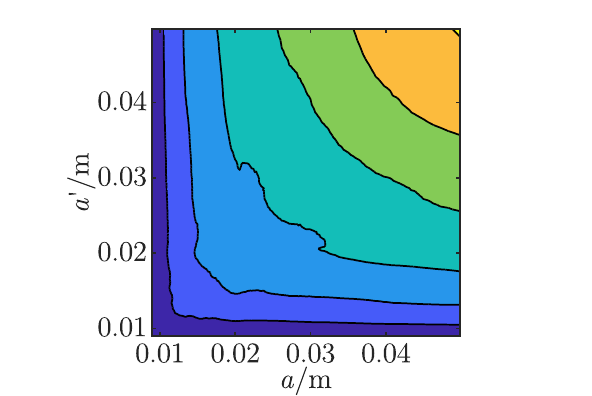} & \includegraphics*[scale=1.0]{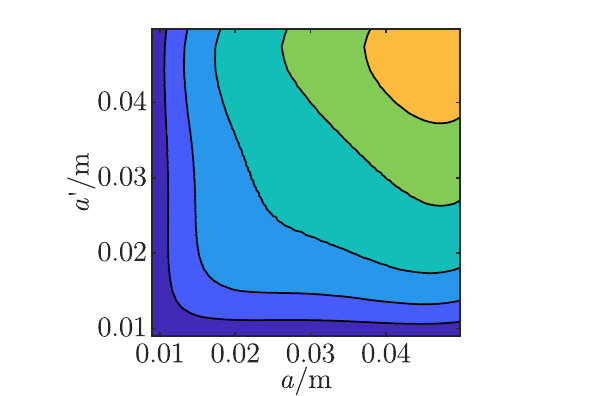} \\
	(a) & (b) \\
	 \\
	\includegraphics*[scale=1.0]{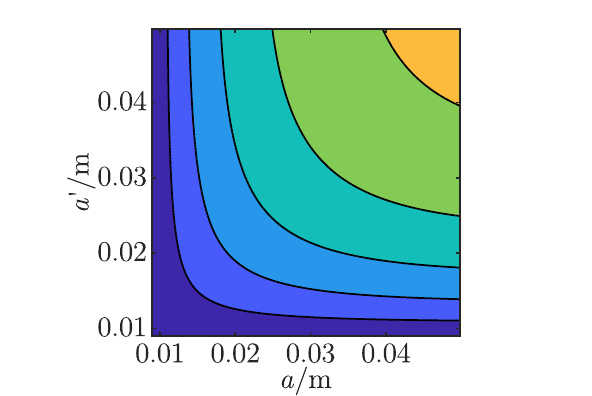} & \includegraphics*[scale=1.0]{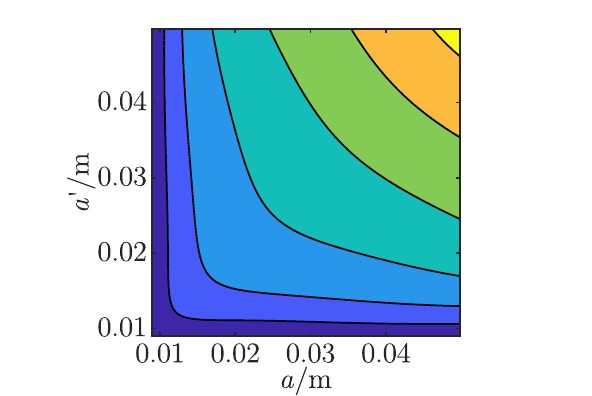} \\
	(c) & (d) \\
	\end{tabular}
  \caption{Comparison of (co-)variances. (a) Computed covariance matrix. (b) Polynomial covariance function with $q=4$. (c) Physics-informed covariance function with $\alpha=2.9$. (d) Physics-informed covariance function with $\alpha=\{2.6,2.8,3.0,3.2\}$.}
  \label{fig:CompareKernels}
\end{figure}

\noindent Figure\,\ref{fig:CompareVirkPred} shows that using different basis functions influences the predictions, see the curvature of the mean towards the end in Figures\,\ref{fig:CompareVirkPred}c and \ref{fig:CompareVirkPred}d. The figure depicts that the polynomial basis functions do not represent the crack growth trajectory as precisely as the physics-informed basis functions. This effect is particularly apparent towards the end of a prediction series, see RMSE half and MAPE half in Table\,\ref{tb:physicResults}. The reason for this might be that in the regions of larger crack lengths, the polynomial covariance function (Figure\,\ref{fig:CompareKernels}b) is significantly different from the computed covariance matrix (Figure\,\ref{fig:CompareKernels}a).

The effect of using several exponents can be observed by comparing Figures\,\ref{fig:CompareVirkPred}a and \ref{fig:CompareVirkPred}b to Figures\,\ref{fig:CompareVirkPred}e and \ref{fig:CompareVirkPred}f. Towards the end, the credible region predicted with several basis functions $\alpha=\{2.6,2.8,3.0,3.2\}$ becomes wider than the credible region predicted with only one $\alpha=2.9$. Moreover, the optimized observation error is smaller for $\alpha=\{2.6,2.8,3.0,3.2\}$. Using additional exponents further improves the predictions, see Table\,\ref{tb:physicResults}. The reason for this might be that not only $C$ but also the material parameter $\alpha$ scatters over different trajectories. The effect of using several values for $\alpha$ can also be seen in Figure\,\ref{fig:CompareKernels}: the covariance function shown in Figure\,\ref{fig:CompareKernels}d is closer to the computed covariance matrix (Figure\,\ref{fig:CompareKernels}a) than the covariance function for $\alpha=2.9$ in Figure\,\ref{fig:CompareKernels}c. For physics-based models also other methods can be used. However, these are restricted to data sets that do have problem-specific governing equations. The method presented in the paper can work with both.


\section{Conclusions}
\noindent The present paper shows that incorporating previous data improves accuracy significantly. Two methods (GPM prev. \& IGPM) on how to incorporate previous data into Gaussian processes are presented.
One way for incorporating previous data into Gaussian processes is to train a predefined GPM by maximizing the sum of the trajectory's log marginal likelihoods. However, optimization tends to be numerically expensive, and due to generally missing convexity, the result may not be globally optimal.
By contrast, following the IGPM method, only a linear equation system has to be solved, which reduces the computation effort significantly. Additionally, it allows the user to specify which basis functions best represent the system output rather than implicitly making assumptions about the system output by choosing a mean and a covariance function. The IGPM thus bypasses the challenge of choosing a predefined GPM a priori and opens the way to physics-informed Gaussian processes, which further increases accuracy, see Section\,\ref{sec:physicsResults}.

Additionally, we recognized a useful property for using GPMs with fixed parameters in the context of PHM: the variance of the posterior distribution is known over the entire range of $x$ once the GPM, the observation error, and the measurement locations $\mathbf{x}$, e.g., number of cycles, are determined, see Section\,\ref{sec:polyDiscuss}.

In summary, the results of the paper show that (1) taking previous data into account increases accuracy and look-ahead time significantly, (2) using physics-informed basis functions further increases accuracy, and (3) deriving the mean and the covariance function by using basis functions remarkably decreases computational effort.

\section*{Acknowledgment}
\noindent We thank Dr.\,Mario\,Teixeira\,Parente, Jülich Centre for Neutron Science, for the enriching discussions and his valuable feedback.

This research was conducted as part of the Strubatex research project in the national aeronautical research program LuFo V and was funded by the Federal Ministry for Economic Affairs and Energy based on a decision by the German Bundestag. The authors declare no conflict of interest.

\begin{figure}[h]
  \centering
  \includegraphics*[width=0.3\linewidth]{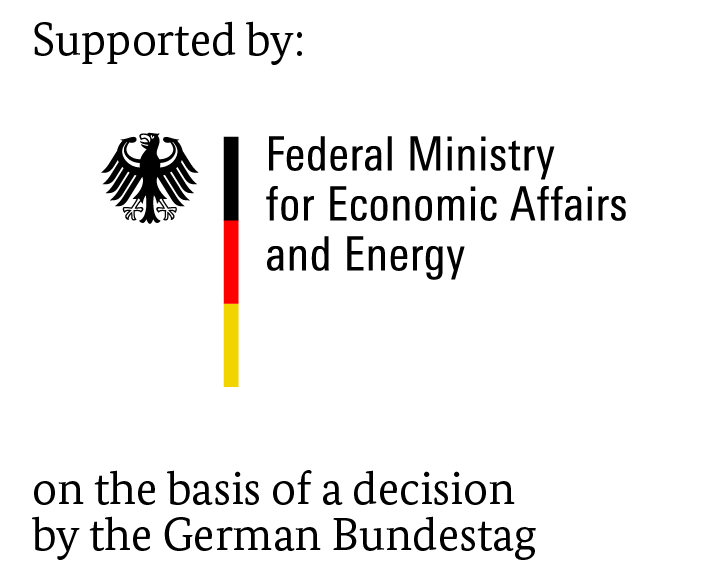}
  \label{fig:BMWi}
\end{figure}


\section*{Appendix}

\noindent Referring to Equations \ref{eq:meanfunction} and \ref{eq:covariancefunction}, we have the following derivations. The mean function $m(x)$ is computed as
\begin{equation} \label{eq:meanDerived}
\begin{split}
	m(x) &= \mathbb{E}[f(x)] \\
	&= \mathbb{E}[\bm{\phi}(x)^\top \bm{\beta}] \\
	&= \bm{\phi}(x)^\top \mathbb{E}[\bm{\beta}] \\
	&= \bm{\phi}(x)^\top \bm{\mu_\beta} \\
	&\approx \bm{\phi}(x)^\top \bm{\hat{\mu}_\beta}.
\end{split}
\end{equation}
Moreover, the covariance function $k(x,x')$ reads
\begin{equation} \label{eq:covDerived}
\begin{split}
	k(x,x') &= \mathrm{cov}[f(x),f(x')] = \mathbb{E}[( f(x) - \mathbb{E}[f(x)] ) ( f(x') - \mathbb{E}[f(x')] )] \\
	&= \mathbb{E}[( f(x) - m(x) ) ( f(x') - m(x') )] \\
	&= \mathbb{E}[f(x)f(x') - f(x)m(x') - m(x)f(x') +  m(x)m(x')] \\
	&= \mathbb{E}[f(x)f(x')] - \mathbb{E}[f(x)m(x')] - \mathbb{E}[m(x)f(x')] +  \mathbb{E}[m(x)m(x')] \\
	&= \mathbb{E}[\bm{\phi}(x)^\top \bm{\beta} \, \bm{\phi}(x')^\top \bm{\beta}] - \mathbb{E}[\bm{\phi}(x)^\top \bm{\beta} \, \bm{\phi}(x')^\top \mathbb{E}[\bm{\beta}]] - \mathbb{E}[\bm{\phi}(x)^\top \mathbb{E}[\bm{\beta}] \, \bm{\phi}(x')^\top \bm{\beta}] +  \mathbb{E}[\bm{\phi}(x)^\top \mathbb{E}[\bm{\beta}] \, \bm{\phi}(x')^\top \mathbb{E}[\bm{\beta}]] \\
	&= \mathbb{E}[\bm{\phi}(x)^\top \bm{\beta} \, \bm{\beta}^\top \bm{\phi}(x')] - \mathbb{E}[\bm{\phi}(x)^\top \bm{\beta} \, \mathbb{E}[\bm{\beta}^\top] \bm{\phi}(x')] - \mathbb{E}[\bm{\phi}(x)^\top \mathbb{E}[\bm{\beta}] \, \bm{\beta}^\top \bm{\phi}(x')] +  \mathbb{E}[\bm{\phi}(x)^\top \mathbb{E}[\bm{\beta}] \, \mathbb{E}[\bm{\beta}^\top] \bm{\phi}(x')] \\
	&= \bm{\phi}(x)^\top \left( \mathbb{E}[\bm{\beta} \, \bm{\beta}^\top] - \mathbb{E}[\bm{\beta}] \, \mathbb{E}[\bm{\beta}^\top] - \mathbb{E}[\bm{\beta}] \, \mathbb{E}[\bm{\beta}^\top] + \mathbb{E}[\bm{\beta}] \, \mathbb{E}[\bm{\beta}^\top] \right) \bm{\phi}(x') \\
	&= \bm{\phi}(x)^\top \left( \mathbb{E}[\bm{\beta} \, \bm{\beta}^\top] - \mathbb{E}[\bm{\beta}] \, \mathbb{E}[\bm{\beta}^\top] \right) \bm{\phi}(x') \\
	&= \bm{\phi}(x)^\top \mathrm{cov}[\bm{\beta}] \bm{\phi}(x') \\
	&= \bm{\phi}(x)^\top \bm{\Sigma_\beta} \bm{\phi}(x') \\
	&\approx \bm{\phi}(x)^\top \bm{\hat{\Sigma}_\beta} \bm{\phi}(x'),
\end{split}
\end{equation}
where $\mathrm{cov}[\bm{\beta}]$ denotes the covariance matrix of the random variable $\bm{\beta}$.


\bibliography{mybibfile}

\begin{thebibliography}{44}
\providecommand{\natexlab}[1]{#1}
\providecommand{\url}[1]{\texttt{#1}}
\expandafter\ifx\csname urlstyle\endcsname\relax
  \providecommand{\doi}[1]{doi: #1}\else
  \providecommand{\doi}{doi: \begingroup \urlstyle{rm}\Url}\fi

\bibitem[Agogino and Goebel(2007)]{agogino2007milling}
A~Agogino and K~Goebel.
\newblock Milling data set.
\newblock \emph{NASA Ames Prognostics Data Repository, BEST Lab: Berkeley, CA,
  USA}, 2007.

\bibitem[An et~al.(2015)An, Kim, and Choi]{an2015practical}
Dawn An, Nam~H Kim, and Joo-Ho Choi.
\newblock Practical options for selecting data-driven or physics-based
  prognostics algorithms with reviews.
\newblock \emph{Reliability Engineering \& System Safety}, 133:\penalty0
  223--236, 2015.

\bibitem[Avendano-Valencia et~al.(2020)Avendano-Valencia, Chatzi, and
  Tcherniak]{avendano2020gaussian}
Luis~David Avendano-Valencia, Eleni~N Chatzi, and Dmitri Tcherniak.
\newblock Gaussian process models for mitigation of operational variability in
  the structural health monitoring of wind turbines.
\newblock \emph{Mechanical Systems and Signal Processing}, 142:\penalty0
  106686, 2020.

\bibitem[Aye and Heyns(2017)]{aye2017integrated}
SA~Aye and PS~Heyns.
\newblock An integrated gaussian process regression for prediction of remaining
  useful life of slow speed bearings based on acoustic emission.
\newblock \emph{Mechanical Systems and Signal Processing}, 84:\penalty0
  485--498, 2017.

\bibitem[Blight and Ott(1975)]{blight1975bayesian}
BJN Blight and L~Ott.
\newblock A bayesian approach to model inadequacy for polynomial regression.
\newblock \emph{Biometrika}, 62\penalty0 (1):\penalty0 79--88, 1975.

\bibitem[Bogdanoff and Kozin(1985)]{bogdanoff1985probabilistic}
John~L Bogdanoff and Frank Kozin.
\newblock Probabilistic models of cumulative damage(book).
\newblock \emph{New York, Wiley-Interscience, 1985, 350 p}, 1985.

\bibitem[Chen et~al.(2020)Chen, Yuan, and Wang]{chen2020line}
Jian Chen, Shenfang Yuan, and Hui Wang.
\newblock On-line updating gaussian process measurement model for crack
  prognosis using the particle filter.
\newblock \emph{Mechanical Systems and Signal Processing}, 140:\penalty0
  106646, 2020.

\bibitem[Do et~al.(2019)Do, Lee, and Nguyen-Xuan]{do2019fast}
Dieu~TT Do, Jaehong Lee, and H~Nguyen-Xuan.
\newblock Fast evaluation of crack growth path using time series forecasting.
\newblock \emph{Engineering Fracture Mechanics}, 218:\penalty0 106567, 2019.

\bibitem[Farrar and Worden(2010)]{farrar2010introduction}
Charles~R Farrar and Keith Worden.
\newblock An introduction to structural health monitoring.
\newblock \emph{New Trends in Vibration Based Structural Health Monitoring},
  pages 1--17, 2010.

\bibitem[Gentile and Galasso(2020)]{gentile2020gaussian}
Roberto Gentile and Carmine Galasso.
\newblock Gaussian process regression for seismic fragility assessment of
  building portfolios.
\newblock \emph{Structural Safety}, 87:\penalty0 101980, 2020.

\bibitem[Gobbato et~al.(2014)Gobbato, Kosmatka, and
  Conte]{gobbato2014recursive}
Maurizio Gobbato, John~B Kosmatka, and Joel~P Conte.
\newblock A recursive bayesian approach for fatigue damage prognosis: An
  experimental validation at the reliability component level.
\newblock \emph{Mechanical Systems and Signal Processing}, 45\penalty0
  (2):\penalty0 448--467, 2014.

\bibitem[Gugulothu et~al.(2017)Gugulothu, TV, Malhotra, Vig, Agarwal, and
  Shroff]{gugulothu2017predicting}
Narendhar Gugulothu, Vishnu TV, Pankaj Malhotra, Lovekesh Vig, Puneet Agarwal,
  and Gautam Shroff.
\newblock Predicting remaining useful life using time series embeddings based
  on recurrent neural networks.
\newblock \emph{arXiv preprint arXiv:1709.01073}, 2017.

\bibitem[Hassani~N. et~al.(2019)Hassani~N., Jin, and Ni]{hassani2019physics}
Seyed M.~Mehdi Hassani~N., Xiaoning Jin, and Jun Ni.
\newblock Physics-based gaussian process for the health monitoring for a
  rolling bearing.
\newblock \emph{Acta Astronautica}, 154:\penalty0 133--139, 2019.

\bibitem[Hong and Zhou(2012)]{hong2012remaining}
Sheng Hong and Zheng Zhou.
\newblock Remaining useful life prognosis of bearing based on gauss process
  regression.
\newblock In \emph{2012 5th International Conference on BioMedical Engineering
  and Informatics}, pages 1575--1579. IEEE, 2012.

\bibitem[Hudak~Jr et~al.(1978)Hudak~Jr, Saxena, Bucci, and
  Malcolm]{hudak1978development}
SJ~Hudak~Jr, A~Saxena, RJ~Bucci, and RC~Malcolm.
\newblock Development of standard methods of testing and analyzing fatigue
  crack growth rate data.
\newblock Technical report, WESTINGHOUSE RESEARCH AND DEVELOPMENT CENTER
  PITTSBURGH PA, 1978.

\bibitem[International(1981)]{astm1981test}
ASTM International.
\newblock Test method for measurement of fatigue crack growth rates.
\newblock \emph{ASTM STP}, 738:\penalty0 340--356, 1981.

\bibitem[Kan et~al.(2015)Kan, Tan, and Mathew]{kan2015review}
Man~Shan Kan, Andy~CC Tan, and Joseph Mathew.
\newblock A review on prognostic techniques for non-stationary and non-linear
  rotating systems.
\newblock \emph{Mechanical Systems and Signal Processing}, 62:\penalty0 1--20,
  2015.

\bibitem[Khelif et~al.(2016)Khelif, Chebel-Morello, Malinowski, Laajili,
  Fnaiech, and Zerhouni]{khelif2016direct}
Racha Khelif, Brigitte Chebel-Morello, Simon Malinowski, Emna Laajili, Farhat
  Fnaiech, and Noureddine Zerhouni.
\newblock Direct remaining useful life estimation based on support vector
  regression.
\newblock \emph{IEEE Transactions on Industrial Electronics}, 64\penalty0
  (3):\penalty0 2276--2285, 2016.

\bibitem[Kong et~al.(2018)Kong, Chen, and Li]{kong2018gaussian}
Dongdong Kong, Yongjie Chen, and Ning Li.
\newblock Gaussian process regression for tool wear prediction.
\newblock \emph{Mechanical Systems and Signal Processing}, 104:\penalty0
  556--574, 2018.

\bibitem[Kwon et~al.(2015)Kwon, Azarian, and Pecht]{kwon2015remaining}
Daeil Kwon, Michael~H Azarian, and Michael Pecht.
\newblock Remaining-life prediction of solder joints using rf impedance
  analysis and gaussian process regression.
\newblock \emph{IEEE Transactions on Components, Packaging and Manufacturing
  Technology}, 5\penalty0 (11):\penalty0 1602--1609, 2015.

\bibitem[Li et~al.(2019)Li, Datta, Chattopadhyay, Iyyer, and
  Phan]{li2019online}
Guoyi Li, Siddhant Datta, Aditi Chattopadhyay, Nagaraja Iyyer, and Nam Phan.
\newblock An online-offline prognosis model for fatigue life prediction under
  biaxial cyclic loading with overloads.
\newblock \emph{Fatigue \& Fracture of Engineering Materials \& Structures},
  42\penalty0 (5):\penalty0 1175--1190, 2019.

\bibitem[Li et~al.(2020)Li, Sadoughi, Hu, and Hu]{li2020hybrid}
Meng Li, Mohammadkazem Sadoughi, Zhen Hu, and Chao Hu.
\newblock A hybrid gaussian process model for system reliability analysis.
\newblock \emph{Reliability Engineering \& System Safety}, 197:\penalty0
  106816, 2020.

\bibitem[Liu et~al.(2013)Liu, Pang, Zhou, Peng, and Pecht]{liu2013prognostics}
Datong Liu, Jingyue Pang, Jianbao Zhou, Yu~Peng, and Michael Pecht.
\newblock Prognostics for state of health estimation of lithium-ion batteries
  based on combination gaussian process functional regression.
\newblock \emph{Microelectronics Reliability}, 53\penalty0 (6):\penalty0
  832--839, 2013.

\bibitem[Lu and Meeker(1993)]{lu1993using}
C~Joseph Lu and William~O Meeker.
\newblock Using degradation measures to estimate a time-to-failure
  distribution.
\newblock \emph{Technometrics}, 35\penalty0 (2):\penalty0 161--174, 1993.

\bibitem[MA and MAO(2019)]{ma2019rotating}
MENG MA and ZHU MAO.
\newblock Rotating machinery prognostics via the fusion of particle filter and
  deep learning.
\newblock \emph{Structural Health Monitoring 2019}, 2019.

\bibitem[Meeker and Escobar(2014)]{meeker2014statistical}
William~Q Meeker and Luis~A Escobar.
\newblock \emph{Statistical methods for reliability data}.
\newblock John Wiley \& Sons, 2014.

\bibitem[Minka and Picard(1997)]{minka1997learning}
Thomas~P Minka and Rosalind~W Picard.
\newblock Learning how to learn is learning with point sets.
\newblock \emph{Unpublished manuscript. Available at http://wwwwhite. media.
  mit. edu/\~{} tpminka/papers/learning. html}, 1997.

\bibitem[Mohanty et~al.(2011)Mohanty, Chattopadhyay, Peralta, and
  Das]{mohanty2011bayesian}
Subhasish Mohanty, Aditi Chattopadhyay, Pedro Peralta, and Santanu Das.
\newblock Bayesian statistic based multivariate gaussian process approach for
  offline/online fatigue crack growth prediction.
\newblock \emph{Experimental Mechanics}, 51\penalty0 (6):\penalty0 833--843,
  2011.

\bibitem[O'Hagan(1978)]{o1978curve}
Anthony O'Hagan.
\newblock Curve fitting and optimal design for prediction.
\newblock \emph{Journal of the Royal Statistical Society: Series B
  (Methodological)}, 40\penalty0 (1):\penalty0 1--24, 1978.

\bibitem[Paris and Erdogan(1963)]{paris1963critical}
Pe~Paris and Fazil Erdogan.
\newblock A critical analysis of crack propagation laws.
\newblock 1963.

\bibitem[Plagemann et~al.(2008)Plagemann, Kersting, and
  Burgard]{plagemann2008nonstationary}
Christian Plagemann, Kristian Kersting, and Wolfram Burgard.
\newblock Nonstationary gaussian process regression using point estimates of
  local smoothness.
\newblock In \emph{Joint European Conference on Machine Learning and Knowledge
  Discovery in Databases}, pages 204--219. Springer, 2008.

\bibitem[Rasmussen and Williams(2008)]{Rasmussen.2008}
Carl~Edward Rasmussen and Christopher K.~I. Williams.
\newblock \emph{Gaussian processes for machine learning}.
\newblock Adaptive Computation and Machine Learning. {MIT Press}, Cambridge,
  Mass., 3. print edition, 2008.
\newblock ISBN 9780262182539.

\bibitem[Saha et~al.(2010)Saha, Saha, Saxena, and Goebel]{saha2010distributed}
Sankalita Saha, Bhaskar Saha, Abhinav Saxena, and Kai Goebel.
\newblock Distributed prognostic health management with gaussian process
  regression.
\newblock In \emph{2010 IEEE Aerospace Conference}, pages 1--8. IEEE, 2010.

\bibitem[Schijve(2009)]{schijve2009fatigue}
Jaap Schijve.
\newblock Fatigue damage in aircraft structures, not wanted, but tolerated?
\newblock \emph{International Journal of Fatigue}, 31\penalty0 (6):\penalty0
  998--1011, 2009.

\bibitem[Sikorska et~al.(2011)Sikorska, Hodkiewicz, and
  Ma]{sikorska2011prognostic}
JZ~Sikorska, Melinda Hodkiewicz, and Lin Ma.
\newblock Prognostic modelling options for remaining useful life estimation by
  industry.
\newblock \emph{Mechanical Systems and Signal Processing}, 25\penalty0
  (5):\penalty0 1803--1836, 2011.

\bibitem[Song et~al.(2018)Song, Jiang, and Jiang]{song2018predict}
Weizhen Song, Zhansi Jiang, and Hui Jiang.
\newblock Predict the fatigue life of crack based on extended finite element
  method and svr.
\newblock In \emph{AIP Conference Proceedings}, volume 1967, page 030024. AIP
  Publishing LLC, 2018.

\bibitem[Su et~al.(2017)Su, Peng, and Hu]{su2017gaussian}
Guoshao Su, Lifeng Peng, and Lihua Hu.
\newblock A gaussian process-based dynamic surrogate model for complex
  engineering structural reliability analysis.
\newblock \emph{Structural Safety}, 68:\penalty0 97--109, 2017.

\bibitem[Tian et~al.(2014)Tian, Azarian, and Pecht]{tian2014anomaly}
Jing Tian, Michael~H Azarian, and Michael Pecht.
\newblock Anomaly detection using self-organizing maps-based k-nearest neighbor
  algorithm.
\newblock In \emph{Proceedings of the European Conference of the Prognostics
  and Health Management Society}. Citeseer, 2014.

\bibitem[Virkler et~al.(1979)Virkler, Hillberry, and
  Goel]{virkler1979statistical}
Dennis~Andrew Virkler, Brnm Hillberry, and PK~Goel.
\newblock The statistical nature of fatigue crack propagation.
\newblock \emph{Journal of Engineering Materials and Technology}, 101\penalty0
  (2):\penalty0 148--153, 1979.

\bibitem[Wang(2016)]{wang2016k}
Dong Wang.
\newblock K-nearest neighbors based methods for identification of different
  gear crack levels under different motor speeds and loads: Revisited.
\newblock \emph{Mechanical Systems and Signal Processing}, 70:\penalty0
  201--208, 2016.

\bibitem[Wang et~al.(2017)Wang, Hu, and Armstrong]{wang2017fatigue}
Wenyi Wang, Weiping Hu, and Nicholas Armstrong.
\newblock Fatigue crack prognosis using bayesian probabilistic modelling.
\newblock \emph{Mechanical Engineering Journal}, 4\penalty0 (5):\penalty0
  16--00702, 2017.

\bibitem[Weber(2017)]{weber2017international}
Ludwig Weber.
\newblock \emph{International civil aviation organization}.
\newblock Kluwer Law International BV, 2017.

\bibitem[Yu(2018)]{yu2018state}
Jianbo Yu.
\newblock State of health prediction of lithium-ion batteries: Multiscale logic
  regression and gaussian process regression ensemble.
\newblock \emph{Reliability Engineering \& System Safety}, 174:\penalty0
  82--95, 2018.

\bibitem[Yuan et~al.(2019)Yuan, Wang, and Chen]{yuan2019pzt}
Shenfang Yuan, Hui Wang, and Jian Chen.
\newblock A pzt based on-line updated guided wave-gaussian process method for
  crack evaluation.
\newblock \emph{IEEE Sensors Journal}, 20\penalty0 (15):\penalty0 8204--8212,
  2019.

\end{thebibliography}

\end{document}